\documentclass[sigconf, nonacm]{acmart}
\usepackage[ruled,vlined]{algorithm2e}
\usepackage{algorithmic} 

\newcommand\vldbdoi{XX.XX/XXX.XX}

\usepackage{url}

\newcommand\vldbavailabilityurl{http://vldb.org/pvldb/format_vol14.html}

\begin{document}
\title{Data Quality Toolkit: Automatic assessment of data quality and remediation for machine learning datasets}


\author{Nitin Gupta$^{1}$, Hima Patel$^{1}$, Shazia Afzal$^{1}$, Naveen Panwar$^{1}$, Ruhi Sharma Mittal$^{1}$, Shanmukha Guttula$^{1}$, Abhinav Jain$^{2}$, Lokesh Nagalapatti$^{3}$, Sameep Mehta$^{1}$, Sandeep Hans$^{1}$, Pranay Lohia$^{1}$, Aniya Aggarwal$^{1}$, Diptikalyan Saha$^{1}$}
\affiliation{%
  \institution{IBM Research, India $^{1}$}
}

\email{{ngupta47,himapatel}@in.ibm.com}


\begin{abstract}
The quality of training data has a huge impact on the efficiency, accuracy and complexity of machine learning tasks. Various tools and techniques are available that assess data quality with respect to general cleaning and profiling checks. However these techniques are not applicable to detect data issues in the context of machine learning tasks, like noisy labels, existence of overlapping classes etc. We attempt to \textit{re-look} at the data quality issues in the context of building a machine learning pipeline and build a tool that can detect, explain and remediate issues in the data, and systematically and automatically capture all the changes applied to the data. We introduce the Data Quality Toolkit for machine learning as a library of some key quality metrics and relevant remediation techniques to analyze and enhance the readiness of structured training datasets for machine learning projects. The toolkit can reduce the turn-around times of data preparation pipelines and streamline the data quality assessment process. Our toolkit is publicly available via IBM API Hub\cite{ibm_apihub} platform, any developer can assess the data quality using the IBM's Data Quality for AI apis\cite{dq_apihub}. Detailed tutorials are also available on IBM Learning Path\cite{ibm_lp}. 


\end{abstract}

\maketitle



\section{Introduction}
\begin{figure*}
  \centering
  \includegraphics[width=\textwidth]{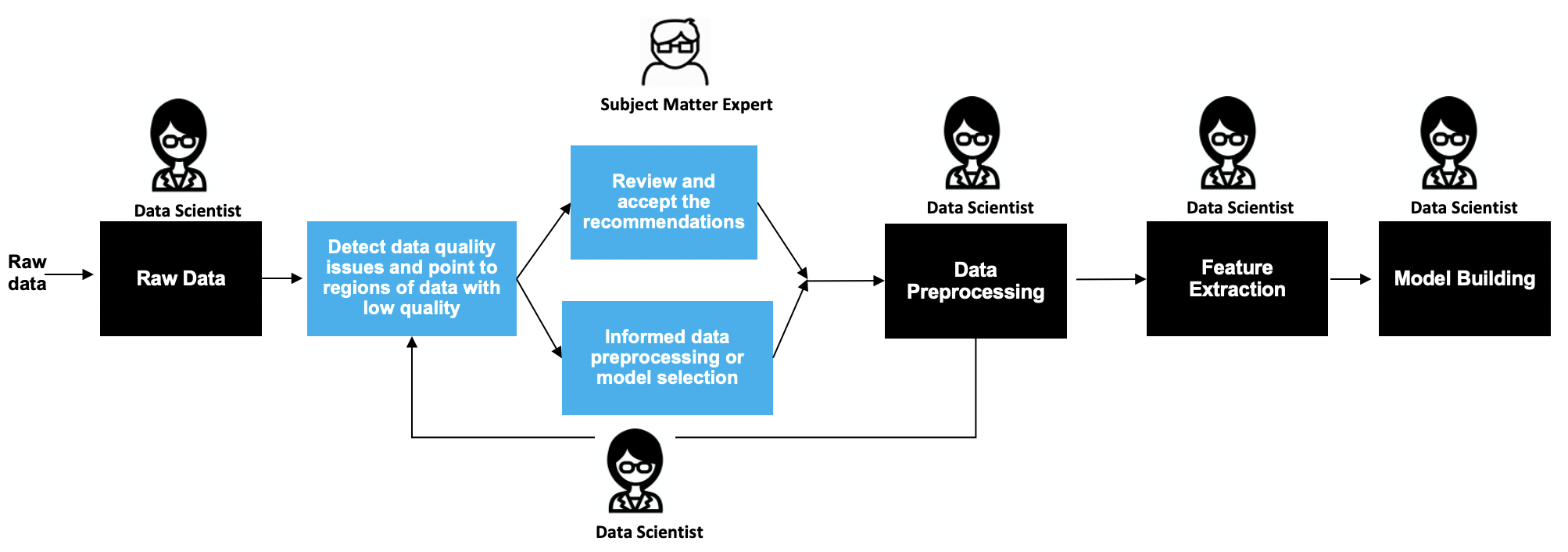}
  \caption{Positioning of the Data Quality Toolkit. Blue boxes represent data readiness toolkit flow as part of a standard data science pipeline (black boxes).}
  \label{fig:dart_positioning}
\end{figure*}

It is well understood from literature that the performance of a machine learning (ML) model is upper bounded by the quality of the data \cite{bandyopadhyay20211st, gupta, 10.1145/3394486.3406477, tut}. While researchers and practitioners have focused on improving the quality of models (such as neural architecture search and automated feature selection), there are limited efforts towards improving the data quality. One of the crucial requirements before consuming datasets for any application is to understand the dataset at hand and failure to do so can result in inaccurate analytics and unreliable decisions. Assessing the quality of the data across intelligently designed metrics and developing corresponding transformation operations to address the quality gaps helps to reduce the effort of a data scientist for iterative debugging of the ML pipeline to improve model performance. Thus, there is a need for tools and algorithms that can reduce the data preparation time. In this report we will highlight the IBM Data Quality Toolkit to systematically measure the quality of data for building machine learning models. Figure \ref{fig:dart_positioning} shows an overall structure of how we position our data quality toolkit (blue boxes) as part of a standard data science pipeline (black boxes). 

The toolkit is designed to serve four primary objectives:
\begin{itemize}
    \item Analyze the quality of data at step $0$ of a data science lifecycle, so that a data scientist can take informed decisions during later stages of model development.
    \item Provide explanations for cases of low quality data by pointing to regions of data responsible for low quality.
    \item Recommend actions to improve the quality of data and provide an easy way to execute the recommendations.
    \item Present an auto generated report that captures the history of operations on the dataset for easy reference to all the changes made to the data, and also to serve for governance and audit usecases. 
\end{itemize}

This toolkit can thus serve as a decision support system to the data scientist/steward and other personas, so that they can take informed decisions. These decisions could be: selecting a model that can take care of identified issues like label noise, dropping features that may not be relevant, deciding that the data is of very low quality and giving feedback to the data acquisition process, etc. 

The data quality toolkit is composed of a set of metrics that quantify the data issues serving as an indicator of how ready the ingested data is for downstream machine learning tasks. Each metric quantifies a specific quality dimension on a scale of 0 to 1. A score of 1 indicates that there are no observed issues for the respective quality aspect. The remediations are expected to improve the quality score by addressing the issues identified. A user can also generate a Data Readiness Report\cite{afzal2020data} to preserve the quality calibration and the various operations performed via the toolkit. The following sections describe the toolkit and the individual features in more detail. Section \ref{section:overview} gives an overview of the toolkit, the general workflow and the various components including the data readiness report in Section \ref{section:Data_Readiness_Report}. Section \ref{section:structured_metrics} lists the various quality metrics and accompanying remediations for structured datasets. Finally we conclude and lay out our plans for future work to address limitations and enhance the current offering in Section \ref{section:future_work}.

\section{Data Quality Toolkit Overview}
\label{section:overview}

Data Quality Toolkit consists of three important components (a) Data Quality Measurement, (b) Data Remediation, and (c) Data Readiness Report, which are described below. 

\subsection{Data Quality Measurement}
Data quality measurement is the most important and largest component of the toolkit. It takes data and evaluates the quality of data based on different quality aspects. Details and explanation of the quality metrics currently included in the toolkit are provided in \autoref{section:structured_metrics}. For each quality metric, this component produces detailed output primarily consisting of a \textit{quality score, explanation, recommended actions and additional details on related analysis}. The quality score quantifies the quality measurement of interest as a real number between 0 to 1. A score of 1 indicates that no issues are identified in the ingested data for the specific metric. A textual explanation is also provided to help users interpret the meaning of the score. Additionally, recommendations are provided to guide users on methods or techniques that can be applied to rectify the issues identified.

The metric output is generated in a human-readable \textit{JSON} format for end-user consumption as well as to communicate with the other components of the toolkit. Specifically, the json is taken as input for follow-up remedial actions by the relevant remediation functions. The quality JSONs are also aggregated and compiled to prepare the Data Readiness Report described in \autoref{section:Data_Readiness_Report}.

\subsection{Data Remediation}
After the data quality analysis, the data remediation component can be invoked to improve and amend the data quality issues identified by the individual quality metrics. Depending on the type of quality measurement, this step may involve a human in the loop process to incorporate human review and feedback before modification or transformation of data. For example, if the label purity metric suggests a new class label for a data point then the data remediation component will confirm with users whether they want to include new class label for data point in updated data or not. After applying all the data remediation functions, the data remediation writes the updated data to the file system. 

\begin{figure*}[t]
  \centering
  \includegraphics[width=1.0\textwidth]{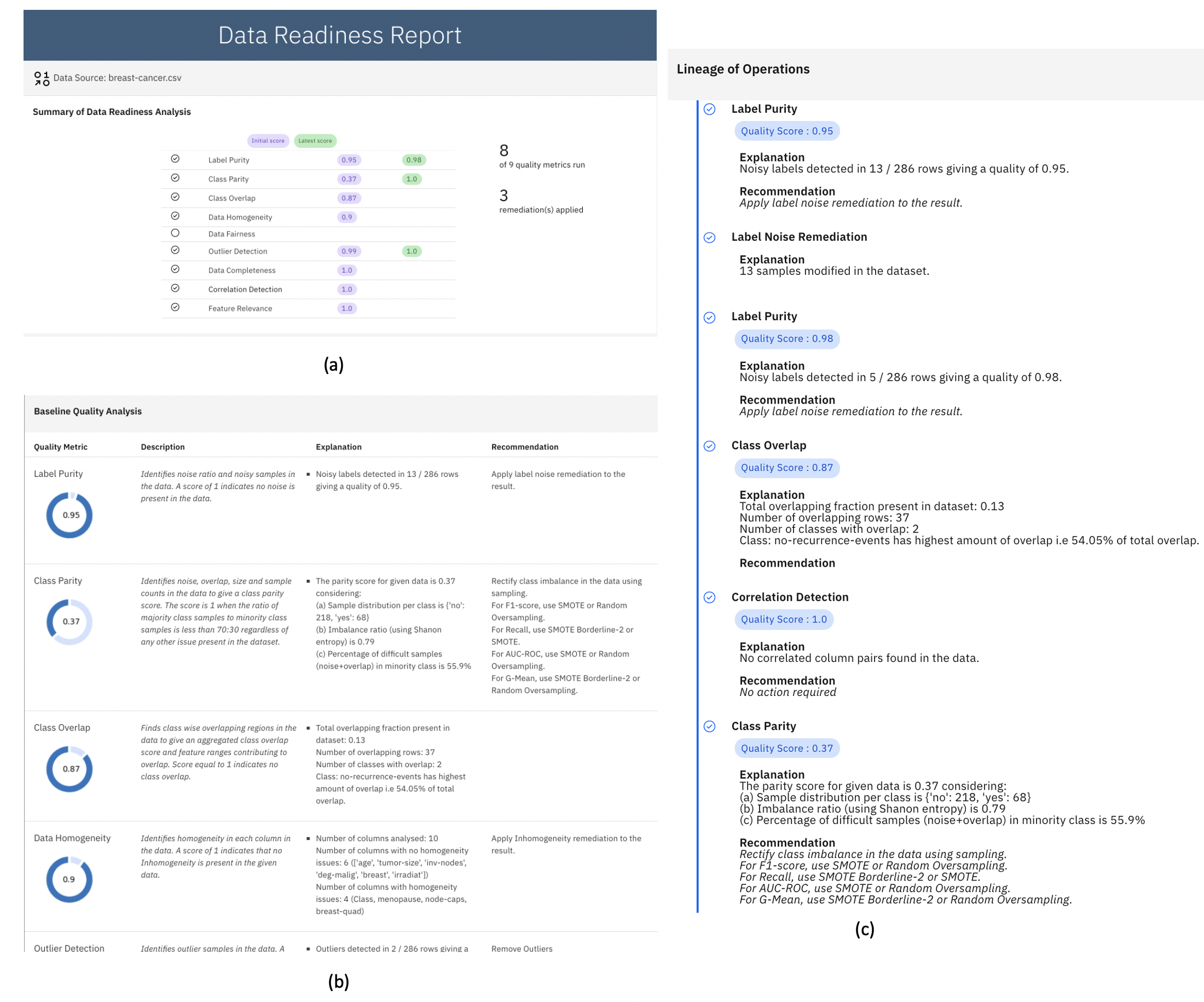}
  \caption{A snapshot of the data readiness report generated after running 8 quality metrics and 3 remediations from the toolkit on a structured dataset. (a) shows summary of data readiness analysis, (b) shows baseline quality analysis on original data, and (c) shows the lineage of operations.}
  \label{fig:structured_report}
\end{figure*}

\subsection{Data Readiness Report}
\label{section:Data_Readiness_Report}
A Data Readiness Report is a shareable asset that serves as a one stop shop of the baseline quality of input dataset as well as a record of operations and remediations performed on it~\cite{afzal2020data}. It is essentially a single artifact for understanding the quality of data and the transformations it has undergone. As mentioned earlier on, data practitioners spend a significant percentage of their time in exploring and tackling various data quality issues. This is often because they have very limited knowledge about the challenges present in incoming data and whether any modifications or changes have been done to it, and if so, by whom.  The data readiness report serves as a comprehensive record of all data properties and quality issues including lineage of all data operations to give a detailed record of how data has evolved. As an accompanying documentation to the data quality toolkit, the report provides data consumers with detailed insights into the quality of input data across the various quality dimensions applied. This apprises data practitioners about the challenges in the data upfront and can potentially reduce iterative data exploration. For more details, please refer to \cite{afzal2020data}.

It is important to note that the data readiness report is not a one time static summary but more like a dynamic, evolving document that can be generated at any time during interaction with the toolkit. When invoked at any point of time, the report generation module would compile and aggregate the quality and remediation outputs to create the most current version of the data readiness report to reflect the operations conducted until time of invocation. Figure \ref{fig:structured_report} show sample of a data readiness report obtained after running the quality metrics and remediations from the toolkit on the UCI breast-cancer dataset~\cite{Dua:2019}.

\section{Structured Metrics}
\label{section:structured_metrics}
Structured data is data that conforms to a consistent format, typically following some data model specification. It consists of rows and columns in a tabular format and is one of the most commonly available data types. Standard solutions that work on structured datasets, such as AutoML, which focus on automation in the steps of feature engineering and model building, do not take into consideration the quality of input data for analysis.  Our toolkit offers a set of standardized metrics to measure the quality of structured data from the perspective of machine learning. Each metric has a quality measurement and a remediation component. The former analyses the quality dimension and quantifies it in the form of a quality score. The quality score is a real number in the range of 0 to 1 where a score of 1 indicates ideal data quality. The remediation component takes the quality assessment and human input, if required, to address the data issues diagnosed in the quality assessment. The following sections describe the metrics currently included in the toolkit for structured data.

    

\subsection{Class Overlap} 
\begin{figure*}[t]
    \centering
    \includegraphics[width=\linewidth]{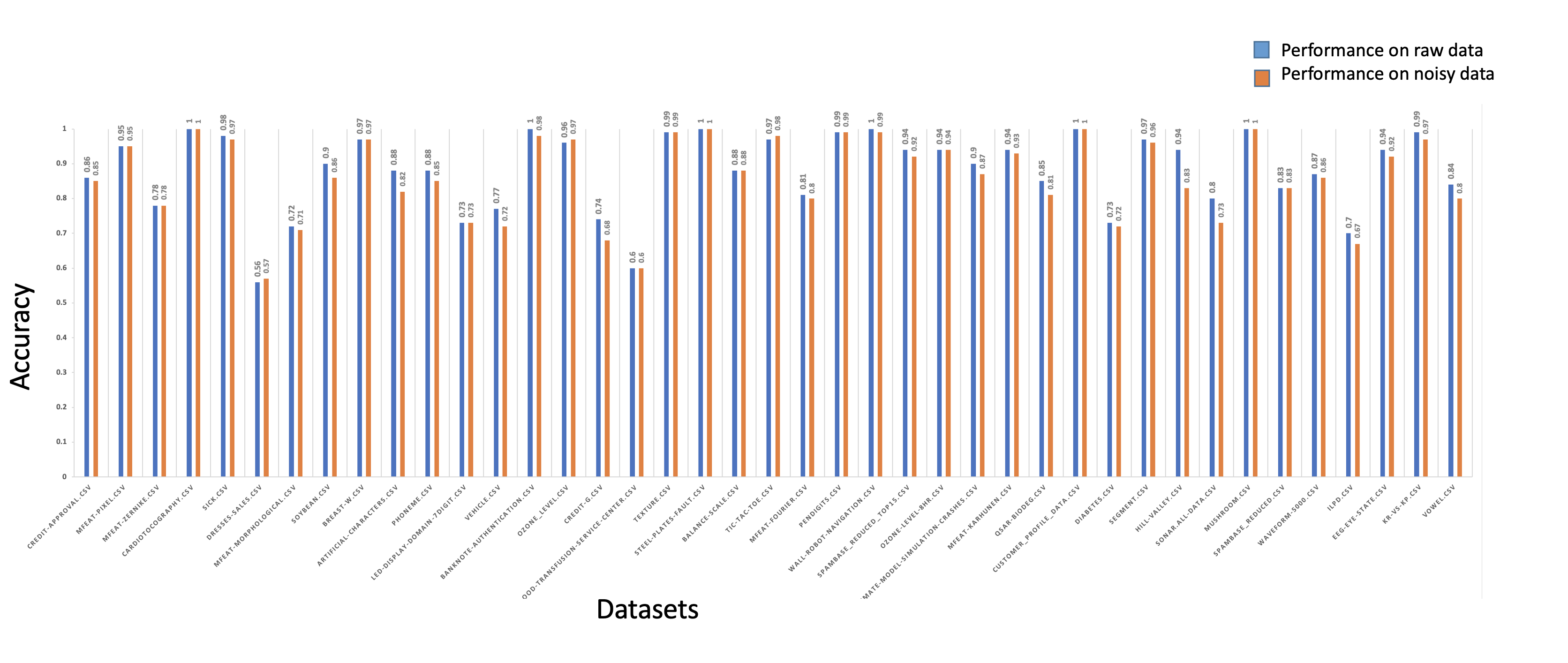}
    \caption{Impact of overlap points in datasets on AutoML classifier accuracy.}
    \label{fig:ov_cl}
\end{figure*}

Overlapping regions among the classes can cause ML classifiers to either mis-classify points or less confident about predicted class in these regions. These effects can be seen in Figure \ref{fig:ov_cl}, as we performed SOA Auto AI classifier \cite{liu2020admm} on datasets, selected from UCI \cite{UCI} and Kaggle \cite{kaggle} repositories. We use $3$ fold cross-validation wherein each fold we added $20\%$ overlap points. We can clearly observe a drop in the performance of classifier after adding overlap points in the training splits. For 18 datasets out of 25, there is decrease in performance by more than $1\%$. There are 12 datasets where decrease in performance is high (greater than $4\%$) and maximum drop observed for these datasets is $15\%$.

We proposed an overlap detection metric which analyzes the dataset to find the data points that reside in the overlapping region of the data space. This includes data points which are close to each other but belong to different classes as well as data points which lies closer to or other side of the class boundary.

The quality measurement of this metric  identifies class wise overlapping regions in the data to give an aggregated class overlap score and feature ranges contributing to overlap. Score equal to 1 indicates no class overlap. The remediation component presently uses manual transformation of features to reduce overlap across the classes.

We validate the performance of proposed algorithm by utilizing the precision and recall plots. We use the approach followed in \cite{8746159} to introduce $20\%$ overlap points in the cleaned data, and then run our algorithm on top of it to measure the effectiveness of the algorithm. We used 19 datasets from UCI \cite{UCI} and Kaggle \cite{kaggle} repositories. We use $3$ fold cross-validation. On an average, the precision of the overlap detection algorithm is above $.80$ and recall is above $.90$.

\begin{figure}[t]
    \centering
    \includegraphics[width=\linewidth]{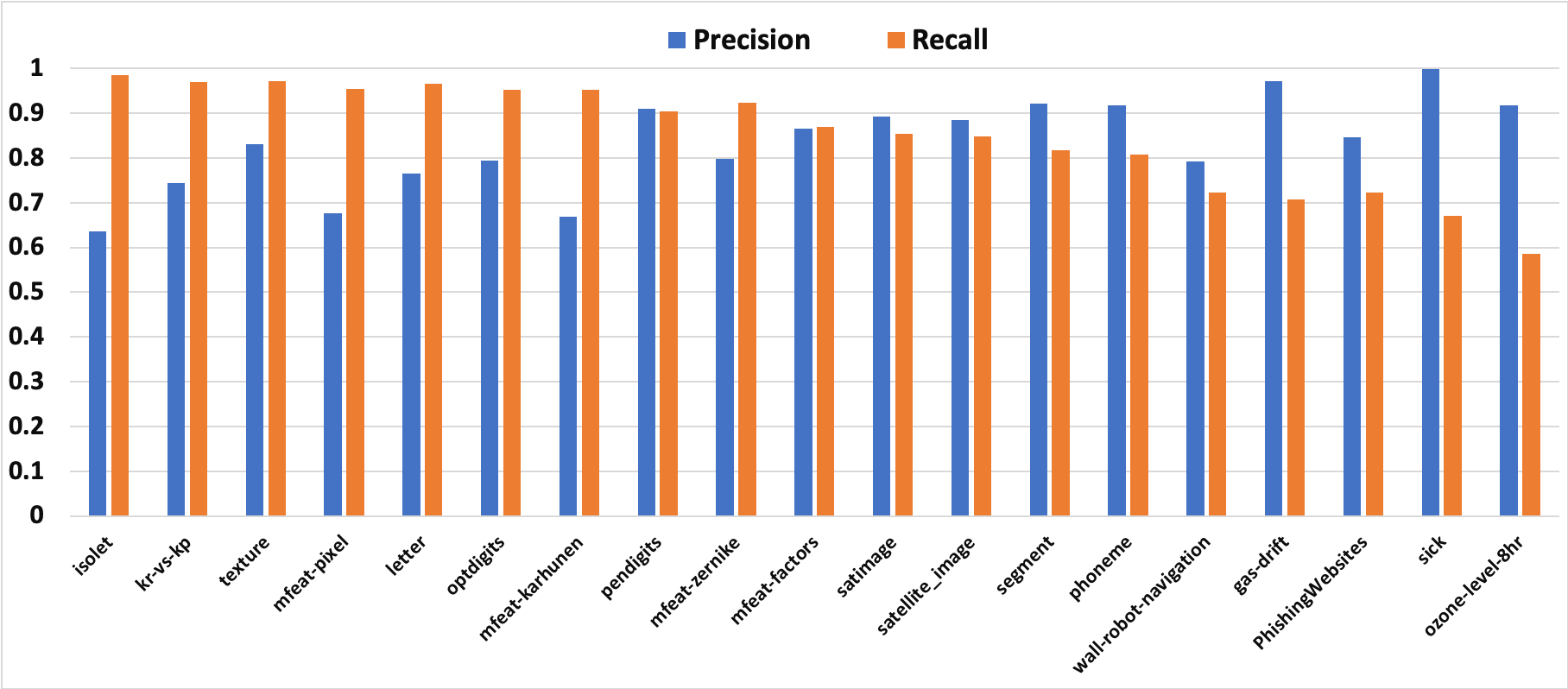}
    \vspace{-2.5mm}
    \caption{Precision and Recall of Overlap Detection Algorithm on 19 datasets (after inducing 30\% overlap points).}
    \label{fig:overla}
    \vspace{-2.5mm}
\end{figure}

\begin{figure*}[t]
    \centering
    \includegraphics[width=\linewidth]{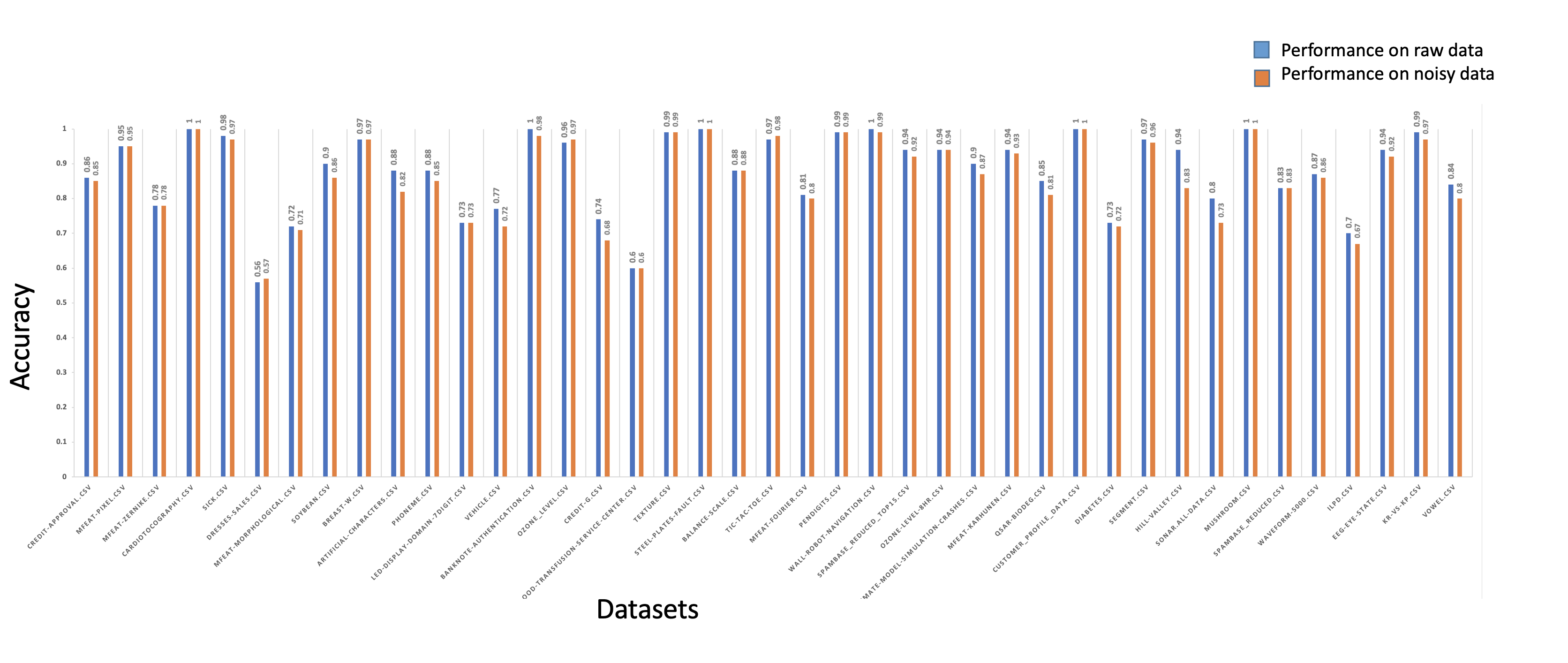}
    \caption{Impact of noisy datasets on AutoML classifier accuracy.}
    \label{fig:lp_cl}
\end{figure*}

\subsection{Label Purity} Most of the real-world datasets generated or annotated have some inconsistent/noisy labels. The label purity metric helps data scientists and end users to identify the label errors or inconsistencies and correct them to model data better. Training data with noisy or inconsistent annotations or labels  \cite{northcutt2019confident} has potentially impact on data science pipeline. For instance, label noise can lead to a decrease in model accuracy, an increase in model complexity, and an increase in the number of training samples required \cite{brodley1999identifying}. Figure \ref{fig:lp_cl} shows the effect of inducing $10\%$ random noise on the performance of SOA Auto AI classifier \cite{liu2020admm} on 41 datasets from UCI \cite{UCI} and Kaggle \cite{kaggle} repositories. We use $3$ fold cross-validation. We can clearly observe a drop in the performance of classifiers after noise is induced. For 25 datasets out of 41, there is decrease in performance by more than $1\%$. There are 8 datasets where decrease in performance is high (greater than $4\%$) and maximum drop observed for these datasets is $11\%$.

\begin{figure*}[t]
  \centering
  \includegraphics[width=1.0\textwidth]{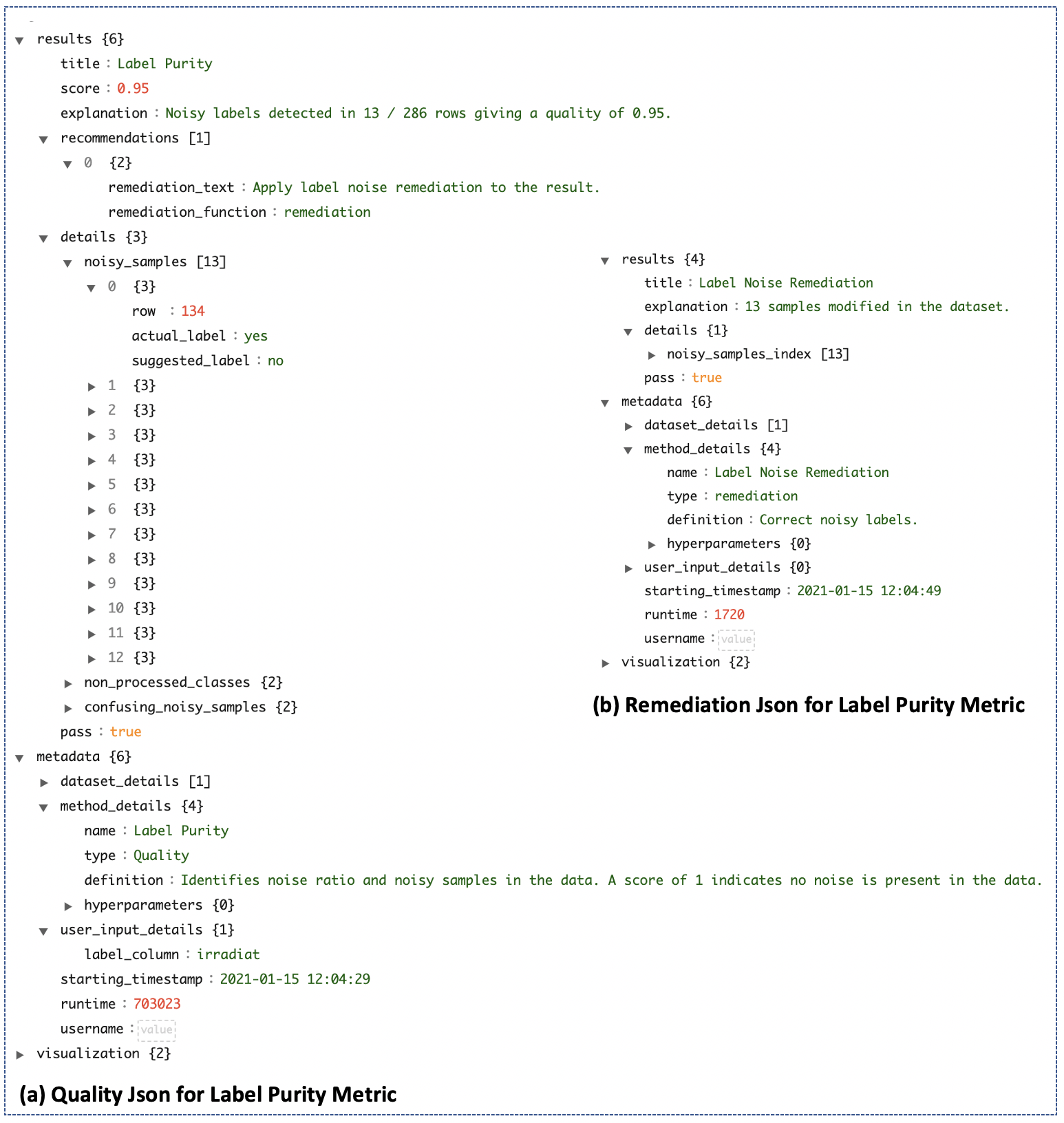}
  \caption{Samples of quality and remediation jsons for Label Purity metric.}
  \label{fig:vis_code}
\end{figure*}

The quality component of this metric identifies noise ratio and the number of noisy samples in the data. A score of 1 indicates that no noise is present in the data. The quality output gives the indices of the noisy samples highlighting the original labels as well as suggesting the correct labels for each (see JSON returned by the tool in Fig. \ref{fig:vis_code}). These are processed by the corresponding remediation component which can either take a human in the loop to review and validate suggested label corrections or trigger an auto remediation mode to automatically apply all suggested corrections by the algorithm.


  \begin{figure*}[t]
    \centering
    \includegraphics[width=\linewidth]{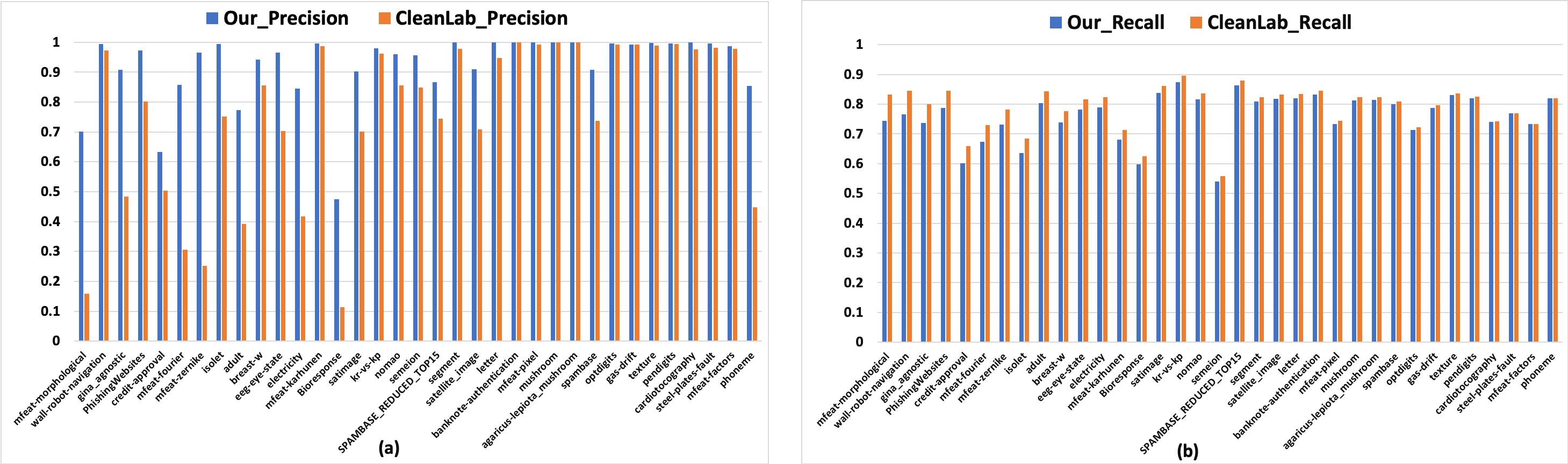}
    \caption{Comparison of (a) Precision and (b) Recall of Label Noise Algorithm with CleanLab Algorithm \cite{northcutt2019confident} on 35 Datasets.}
    \label{fig:labelnoise_prec_recall}
\end{figure*}

We validate the improvement of our proposed algorithm by comparing the precision and recall (see Figure \ref{fig:labelnoise_prec_recall}) with CleanLab as baseline. We demonstrate these results on $35$ datasets selected from UCI \cite{UCI} and Kaggle \cite{kaggle} repositories. We use $3$ fold cross-validation wherein each fold, we introduce $10\%$ random noise per class in the training set only and do not make any changes to the test set. The charts in Figure \ref{fig:labelnoise_prec_recall} show that our algorithm outperforms CleanLab \cite{northcutt2019confident} in terms of precision ($5-15 \%$ improvement). For recall, both the algorithms have similar performance.

 \begin{figure*}[t]
    \centering
    \includegraphics[width=\linewidth]{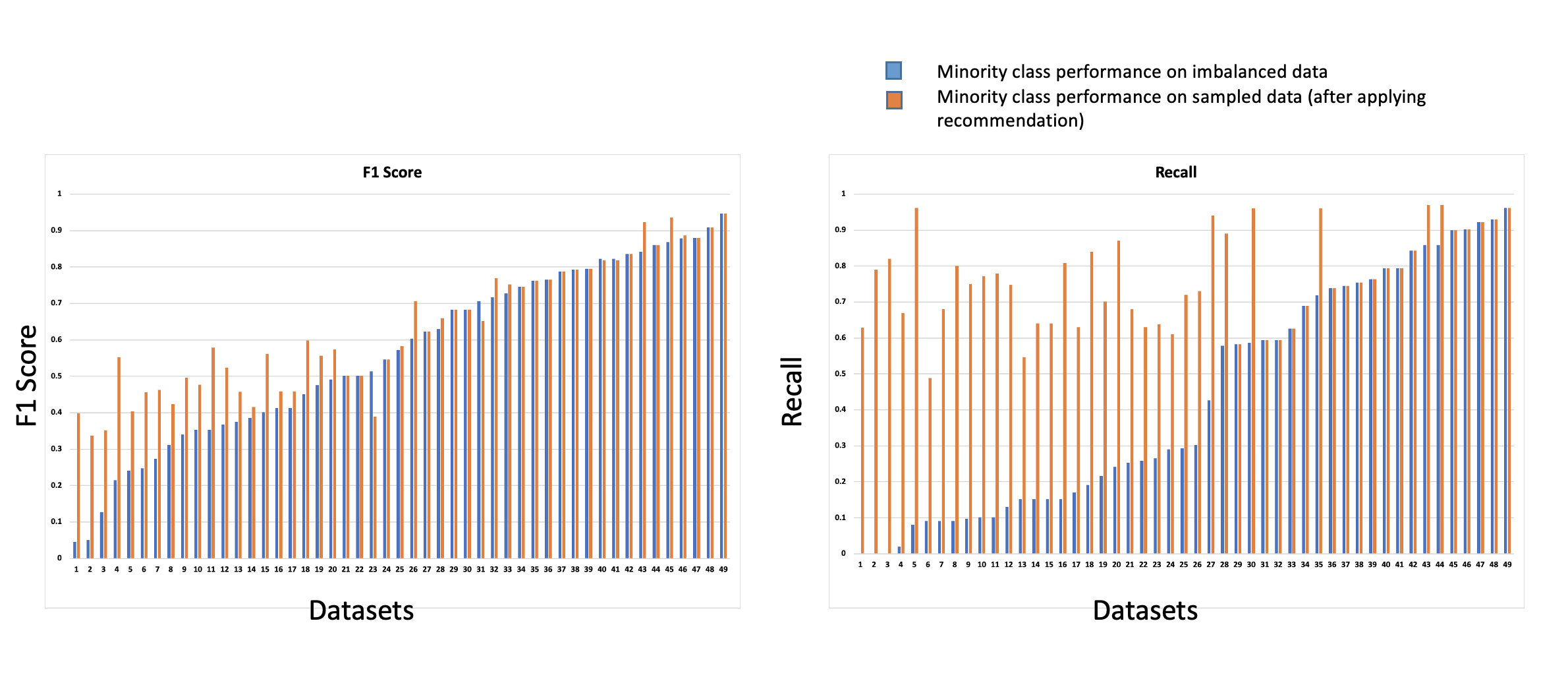}
    \caption{Impact of imbalanced datasets on AutoML classifier minority F1 and Recall score.}
    \label{fig:par_cl}
\end{figure*}
\subsection{Class Parity} 

Imbalanced datasets can cause ML models to be biased towards the majority class. As the imbalance ratio in the dataset increases, the classifier performance may decrease because the learning algorithm becomes more biased towards the majority class. Is imbalance ratio the only cause of performance degradation in learning from imbalanced data? The answer is No. If the imbalance ratio is high, but classes are well-represented and come from non-overlapping distribution, we may obtain good classifier performance. Therefore, there are other factors along with imbalance ratio which deteriorates the performance of imbalanced classification. 
This problem aggravates if minority class has overlapping and noisy points, minority class is divided into sub-concepts and number of data points in minority class are very low. In imbalanced classification, the class which is critical from the domain point of view and has fewer samples is known as minority class whereas other classes are known as majority classes. The major issue with imbalance classification is that the standard algorithms are accuracy driven and assume the dataset to be balanced. When there is an imbalance in classes, priority is given to the majority class compare to the minority class which leads to degradation in minority class performance. 

The proposed class parity metric helps data scientists and end users to analyse class imbalance by looking at several properties of data like imbalance ratio, dataset size, proportion of difficult samples in the extreme minority class and offers remedial re-sampling strategies for balancing data. To show the impact of sampling on the performance of the classifier, we conduct an experiment on 49 imbalanced datasets from keel repository \cite{keel_repository}. The datasets are for binary classification where we take minority class as the positive class. We use 3-fold cross-validation wherein each fold, we over-sample only the training data using recommended sampling. Figure \ref{fig:par_cl} shows F1 and Recall score of minority class on 49 datasets with and without sampling using Auto AI classifier. The positive impact of sampling is evident from the results.


\subsection{Feature Relevance} As the name suggests, this metric helps data scientists and end users to analyze the relative importance of each feature with respect to the target variable or class i.e. high correlation with the target variable, but little correlation with other features. This is more useful for high-dimensional data where the number of features are high and might carry redundant information. The feature relevance metric identifies and ranks the features based on their relevance quantifying the analysis as a feature relevance score. A score of 1 indicates that all features are relevant. The remediation component of the metric removes the less relevant features suggested by the quality metric thus retaining only the features that maximise performance of subsequent classification tasks.

 \begin{figure*}[t]
    \centering
    \includegraphics[width=\linewidth]{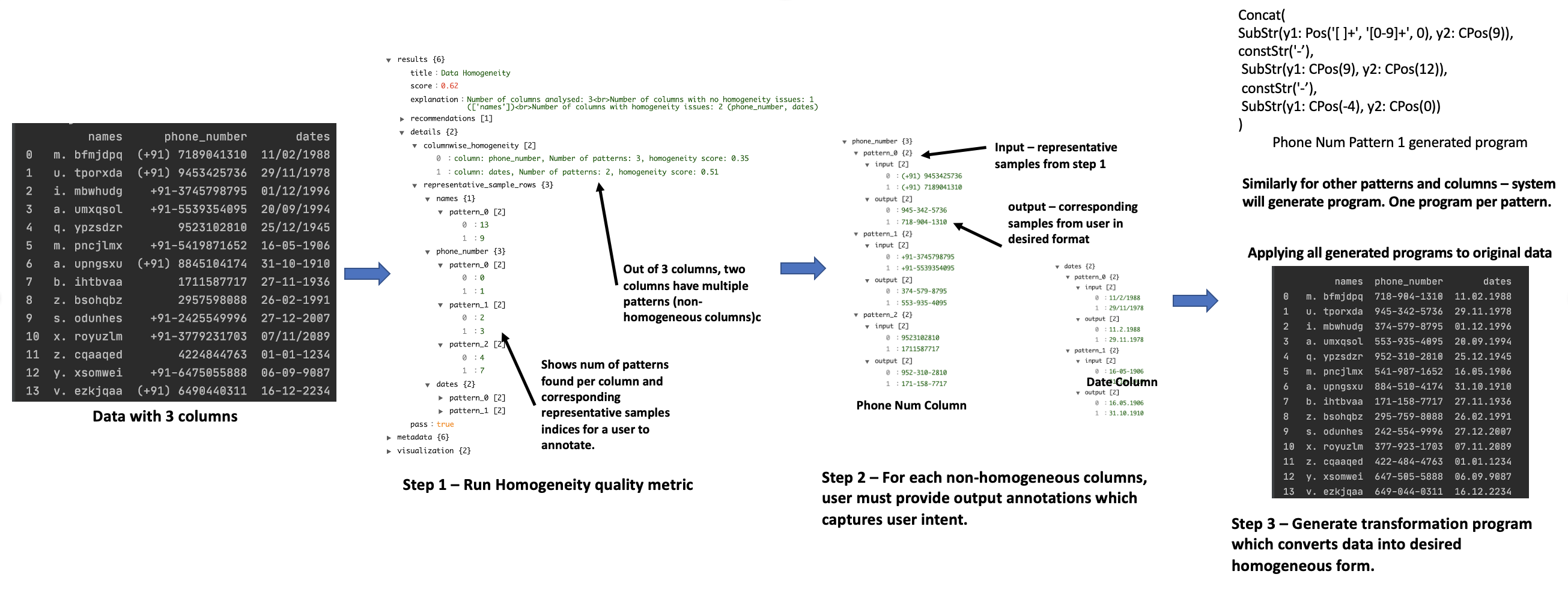}
    \caption{Data Homogeneity Quality and Remediation Illustration.}
    \label{homo}
\end{figure*}
\subsection{Data Homogeneity} Consistent data is necessary for building machine learning models and doing meaningful business analytics. This metric helps data scientists and end users to identify the format inconsistencies in non-numerical columns and fix them by transforming the data into user intended format. For example, multiple data patterns in date column, different formats of telephone number in contact column, etc. The data homogeneity metric identifies homogeneity in each column in the data and returns the representative samples for each column. A score of 1 indicates that data is homogeneous. The corresponding remediation method makes data homogeneous by transforming the samples from specified columns into user intended format. The user is required to provide input in JSON format that captures few examples of expected output for the representative input samples from quality metric. Then system use that information to automatically finds the best possible program to convert source format to target format.

\subsection{Data Fairness} Given the increasing concerns around prevalence of potentially biased information and its impact on model application, identifying and flagging bias in data is a critical aspect of data assessment. Our toolkit provides a data fairness metric that computes group discrimination using disparate impact. It identifies the bias in the dataset and returns the disparate impact score (real number in range set - [0,1]) for sensitive attributes in the given data. A score of 1 stands for highest quality with respect to data fairness metric. The data fairness remediation improves the quality score by mitigating bias in the data by editing feature values to improve group fairness.
 
\subsection{Correlation Detection} 
    A feature or an attribute is correlated when it depends on another attribute or is a cause for another attribute. When a dataset contains attributes that are highly correlated with each other then the problem of multi-collinearity \cite{multicollinearity} arises. If not handled, it leads to over-fitting where the model may do well on a known training set but will fail on an unknown testing set. Therefore, it is important to detect and correct the multi-collinearity issue in the dataset. We use the approach which identifies the correlated numerical columns in the data also computing a score where 1 indicates no correlated columns found in the data. The corresponding remediation fixes the correlation amongst data columns by recommending to drop highest correlated columns with respect to all the columns from each identified correlated column pair.

\subsection{Data Completeness} Large number of missing values in datasets can adversely affect ML models and result in misleading inferences. The data completeness metric in the toolkit helps in detection of missing values and provides imputation mechanisms to model data better. The quality part of the metric identifies the location of missing values in the given data. Similar to other quality scores, a score of 1 indicates no missing values are found in the data. When the score is not perfect, the remediation part of the metric imputes the missing entries using a proposed constraints and association based approach.

The proposed constraints and association based approach for data imputation is based on different data type values and every attribute’s relation with other attributes. 
   
\subsection{Outlier Detection} 
\cite{outlier_definition}, "An outlier is an observation that deviates so much from other observations as to arouse suspicion that it was generated by a different mechanism". The presence of outliers in data increases the misclassification hence degrades the model performance \cite{outlier_impact, agarwal2021comparison}. It is therefore important to detect and remove outliers in the dataset. We use one of the SOA approach for outlier detection. Given its relevance, we include an outlier detection metric in our toolkit to to flag outlier samples and remove them to model data better. For consistency with the other metrics in the toolkit, this metric quantifies the presence  of outliers using an aggregate score between 0 to 1 wherein a score of 1 indicates no outliers found in the data. Exact indices of the outlier samples are given as output so that the complementary remediation can remove these easily.

\subsection{Data Duplicates} Duplicate records are not only unnecessary but can also waste memory, compute time and also leads to the data imbalance. Our toolkit provides a data duplicate detection metric to spot duplicate records. Like other metrics, there is a quality score, between 0 to 1 wherein a score of 1 indicates no outliers found in the data, computed to indicate the degree of duplicates present and a corresponding remediation that removes such records to clean the data. We used pandas ``.duplicated" function to find the duplicates in dataset.

\section{Conclusion and Future Work}
\label{section:future_work}

We have described a data quality toolkit for machine learning aimed at automatic assessment of various data quality dimensions in order to reduce the data preparation time and improve overall training data quality. The toolkit is based on a generic architecture designed to integrate with different database connectors. We have discussed the different quality metrics for structured datasets that are currently offered in the toolkit to help data practitioners understand the challenges in data upfront. The design and usage of the toolkit takes into consideration the range of technical skills in affected data practitioners. To this end, we are exploring ways to incorporate better end-user interactivity and visualization capabilities in the toolkit. 

Finally, the importance of good quality data for building efficient and robust AI systems cannot be emphasised enough. Having tools that assist in determining data quality can improvise the data preparation phase and increase overall productivity and turn-around times for model building and deployment. The data readiness toolkit that we describe is a step in this direction and we plan to expand it in the future to other data modalities like images, speech etc. 



\bibliographystyle{ACM-Reference-Format}
\bibliography{sample}

\end{document}